\documentclass[letterpaper]{article} 
\usepackage{aaai24}  
\usepackage{times}  
\usepackage{helvet}  
\usepackage{courier}  
\usepackage[hyphens]{url}  
\usepackage{graphicx} 
\usepackage{natbib}  
\usepackage{caption} 
\usepackage{algorithm}
\usepackage{algorithmic}

\usepackage{newfloat}
\usepackage{listings}
\DeclareCaptionStyle{ruled}{labelfont=normalfont,labelsep=colon,strut=off} 
\floatstyle{ruled}
\newfloat{listing}{tb}{lst}{}
\floatname{listing}{Listing}
\title{Look Ahead Text Understanding and LLM Stitching}
\author{
		Junlin Julian Jiang\textsuperscript{\rm 1}, 
		Xin Li\textsuperscript{\rm 2}\thanks{Corresponding author.}
}
\affiliations{
    \textsuperscript{\rm 1}Piedmont High School, Piedmont, CA, USA \\

    \textsuperscript{\rm 2}College of Business, City University of Hong Kong, Hong Kong, China \\
    junlinjjiang@gmail.com,  xin.li.phd@gmail.com
}

\usepackage{bibentry}

\begin{document}

\maketitle

\begin{abstract}
This paper proposes a look ahead text understanding problem with look ahead section identification (LASI) as an example. This problem may appear in generative AI as well as human interactions, where we want to understand the direction of a developing text or conversation. We tackle the problem using transformer-based LLMs. We show that LASI is more challenging than classic section identification (SI). We argue that both bidirectional contextual information (e.g., BERT) and unidirectional predictive ability (e.g., GPT) will benefit the task. We propose two approaches to stitch together BERT and GPT. Experiments show that our approach outperforms the established models, especially when there is noise in the text (which is often the case for developing text in generative AI). Our paper sheds light on other look ahead text understanding tasks that are important to social media, such as look ahead sentiment classification, and points out the opportunities to leverage pre-trained LLMs through stitching. 
\end{abstract}

\section{Introduction}

Text understanding problems are well studied in artificial intelligence, with various tasks being proposed to reduce humans' cognitive load. The development of social media and generative AI brings a new challenge to text understanding: sometimes, we want to “understand” a text before it is written or “generated.” In conversations, humans do this often to predict the topic or sentiment of the coming text and prepare a response. This study aims to tackle this problem using section identification (SI) as an example application.

SI is a task where the contents of a document are labeled and separated into a structure based on their semantics \cite{zhou_feature_2020}. For example, scientific papers often contain introductions, related works, methods, results, and conclusions. Web content, patents, technical documents, and financial documents also have their standard formats. Due to the large degree of commonality of document structures, SI provides us with a more structured text understanding task than other tasks, such as topic classification. Meanwhile, document structure plays a significant role in human perception. By providing the structure, we can guide human attention to important components in the text and thus reduce the effort involved in reading. It has also been shown that a document's alignment with their expected structures will increase the audience's trust in the document \cite{rowley_understanding_2013}. 

In practice, researchers often conduct SI on finished documents. We argue that the SI task could go beyond finished documents to developing texts. For example, when AI responds to humans on social media or assists humans in writing long documents, structure analysis and SI can help provide more accurate models (parameters) fitting the next sentence to be generated.\footnote{Long documents with different sections can also be generated by one large model. This paper refers to applications that care about the document structure more.} By transiting across models, AI may generate more accurate content for social media interactions and document co-writing. Aligning with this need, we study a look ahead section identification (LASI) task (Figure 1). 
 
\begin{figure}[t]
\centering
\includegraphics[width=\columnwidth]{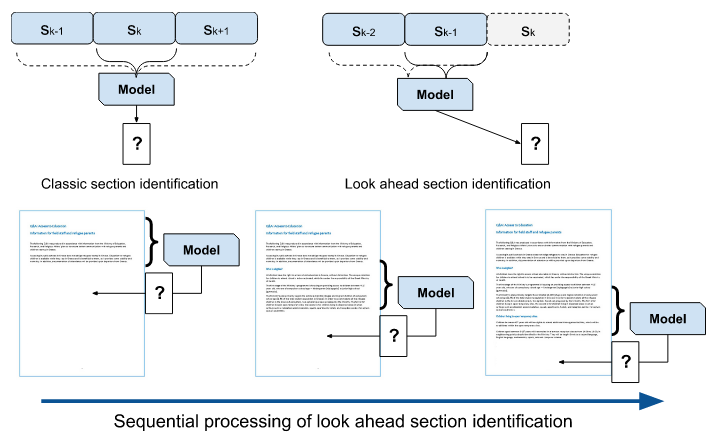} 
\caption{Look Ahead Section Identification}
\label{fig1}
\end{figure}

In this paper, we tackle the LASI problem using transformer-based large language models (LLMs) \cite{vaswani_attention_2017}. We show that LASI is more challenging than the classic SI problem. Given the problem's characteristics, we argue that both bidirectional context information (e.g., captured by BERT) and unidirectional prediction ability (e.g., of GPT) will facilitate the task. We propose stitching GPT and BERT through adjacent sentence pairs (using loss functions or attention modules) to drift GPT output to the space of BERT. Experiments show that this approach achieves good results, especially when there is noise in the text, which is common in generative AI and social media conversations. 

Our contributions are twofold: (i) We highlight the importance of look ahead text understanding through an example of LASI. In a broad sense, the LASI problem has a similar setup to look ahead sentiment classification or look ahead topic classification, both of which have more applications than LASI in social media. By using the LASI example, this paper opens a venue for exploring other look ahead tasks with practical implications in long text generation and social media text mining. (ii) We propose LLM stitching approaches to solve the LASI problem. This approach can be used to leverage LLMs pre-trained on different corpora or for different purposes in one task. This could be a lightweight way to leverage prior efforts on pre-trained LLMs. 

\section{Overview}

\subsection{Problem Setting}

As we will further elaborate in the related works section, earlier SI studies often used abstracts for experiments where sentences are classified into different section labels. Here, we assume a document the $D$ with a sequence of sentences $\{s_k\}$. As illustrated in Figure 1, in a classic setting, one needs to build a classifier $s_k \longrightarrow l_k$ with $l_k \in L$ on a set of section labels. Under this setup, information used for classification mainly consists of linguistic and semantic patterns of $s_k$. 

The assumption of having document $D$ finished before SI also allows us to consider the context information before and after $s_k$ during SI. For example, in the BiLSTM+CRF setup, the CRF model captures inter-sentence dependencies \cite{jin_hierarchical_2018}. In this setup, the problem can be framed as looking for a classifier $\{\cdots s_{k-1}, s_k, s_{k+1} \cdots \} \longrightarrow l_k$, where one or more sentences from the context can be used. 

In this paper, we propose a different setup where document $D$ is not fully developed when conducting SI. In text generation applications or when developing conversations on social media, SI can be performed gradually (sentence by sentence or paragraph by paragraph). In this situation, the task of understanding the coming text (which has not been written yet) based on finished content is a task of LASI.This problem can be framed as a classifier $\{\cdots s_{k-2}, s_{k-1}\} \longrightarrow l_k$, where one or more sentences before $s_k$ can be used to predict its label $l_k$, even though $s_k$ has not yet been written. 

\begin{figure}[t]
\centering
\includegraphics[width=\columnwidth]{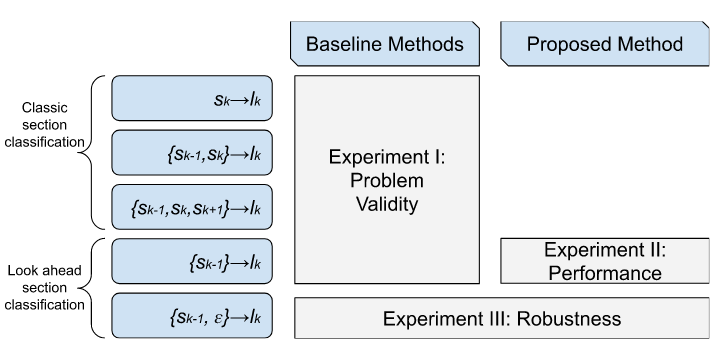} 
\caption{Overview of the Study}
\label{fig2}
\end{figure}

\subsection{Research Framework}

Figure \ref{fig2} illustrates the process of our study, which has three sets of experiments. In the first experiment, we explore the difficulty of LASI compared to classic SI. Aligning with our problem setting elaborated above, we compare setups using different sentences to predict the label of $s_k$. 

In the second experiment, we examine the performance of multiple baseline models and our proposed model in tackling LASI, which we elaborate on in the next section. 

In the third experiment, we further tweak the LASI problem by introducing noise $\epsilon$ into the task. We consider scenarios that may appear in a generative AI setting where the developing document $D$ is imperfect and assess whether our proposed methods are robust to noise.

\section{Methods}

\subsection{BERT vs. GPT}

In existing SI studies, BERT is the most popular large language model. This may be because BERT can capture context information of the entire input and suits semantic classification. 

However, in LASI, our goal is to “understand” the next sentence, which has not yet appeared. BERT has limited power to predict future content. We conjecture that the GPT model may help in this task. GPT is a transformer decoder designed for autoregressive inference, which can predict the next token based on the previous tokens. 

\begin{figure}[b]
\centering
\includegraphics[width=0.7\columnwidth]{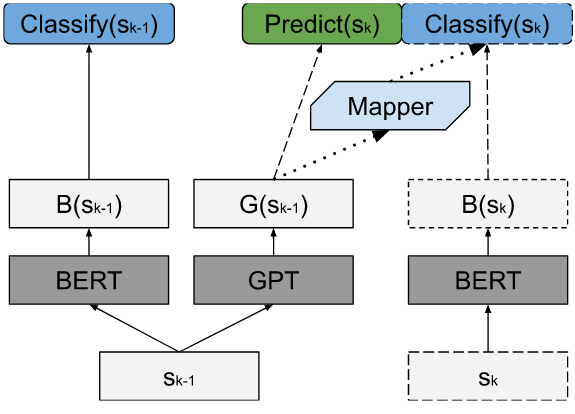} 
\caption{BERT and GPT Representation}
\label{fig3}
\end{figure}

As illustrated in Figure \ref{fig3}, in SI, BERT can classify $s_{k-1}$. If $s_k$ exists, BERT can also classify $s_k$. GPT’s ability to predict the next token (and the next sentence $s_k$) based on $s_{k-1}$ means it captures information closer to $s_k$. Thus, GPT’s output from $s_{k-1}$ may have the potential to classify $s_k$. The question is how to use this information.
 
A naive solution is to directly use GPT output on $s_{k-1}$ for classification. We explore this solution as a baseline. However, what interests us more is how to combine BERT with GPT to tackle the LASI task, which we elaborate on below.

\subsection{Straightforward Methods}

\subsubsection{Generation-Classification.}

A straightforward solution of LASI is to generate the next sentence using GPT and then feed it to BERT for classification, which we refer to as BERT(GPT). If GPT's prediction is perfect, we basically get the next sentence that will appear, and LASI is reduced to the SI task.

\subsubsection{Direct Combination.}

Another option to combine the two models is to directly concatenate the embeddings of BERT and GPT before feeding them to the classification head. We refer to this approach as GPT+BERT. For this model, the classification head is twice the size of the original classification head, which needs to be fine-tuned.

\subsubsection{BART.}

One established solution that leverages both BERT and GPT is BART, which is a full transformer model with both encoders (as BERT) and decoders (as GPT). BART shows good performance in various tasks and we include it in this study. It should be noted that the BART model has a different classification head from the BERT and GPT models. We customize its head to be a single-layer linear neural network for a fair comparison with BERT and GPT.

In BART, the final encoder output through all layers of encoders is fed into each decoder layer through multi-head decoder-encoder attention modules. We understand that this design will drift the decoder representation (based on prior tokens) toward the encoder output (based on the context). The BART model potentially combines the power of encoders and decoders by weaving them together with complicated connections. This approach creates a coherent model that needs to be pre-trained. 

\subsection{Model Stitching}

As illustrated in Figure \ref{fig3}, if we believe GPT output on $s_{k-1}$ and BERT output on $s_k$ contain common information that can be used to classify $s_k$, we may be able to make connections between their representations in different spaces. If we can map GPT to the BERT space or drift GPT output toward BERT representation, we may have additional information for LASI. Compared with BART, which weaves encoders with decoders (resulting in a complicated model that needs pre-training), we intend to explore simple solutions that can stitch LLMs together during fine-tuning. 

\begin{figure}[h]
\centering
\includegraphics[width=0.8\columnwidth]{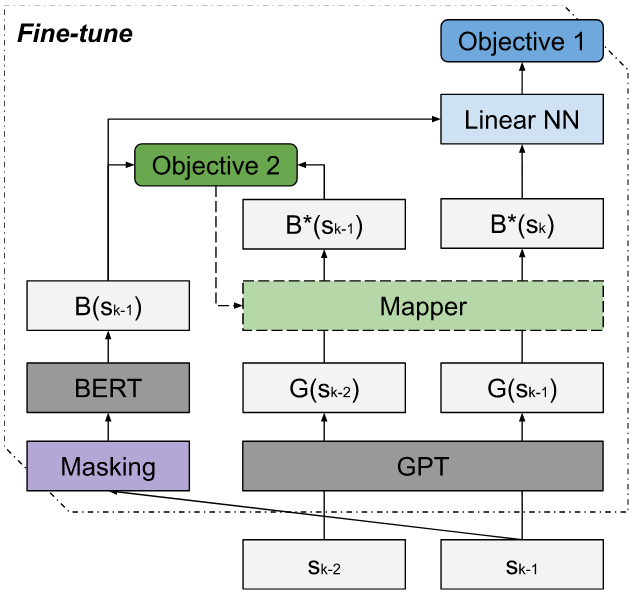} 
\caption{Loss Stitching}
\label{fig4}
\end{figure}

\subsubsection{Mapping of Representation Spaces.}

For sentence $s_{k-1}$, GPT delivers $G(s_{k-1})$ as the output. For sentence $s_k$, BERT delivers $B(s_k)$ as the output. Since we consider them to capture common information on $s_k$, we assume there is a mapping $f()$ between them as follows:
\begin{equation}
B(s_k)=f(G(s_{k-1}))
\end{equation}
If we find the perfect $f()$, we essentially convert GPT on $s_{k-1}$ to BERT on $s_k$, which reduces the LASI problem to a classic SI problem. In reality, we look for approximations, such as a linear mapping: 
\begin{equation}
B(s_k)^*=L \cdot G(s_{k-1})
\end{equation}
where $B(s_k)^*$ can be approximated by $B(s_k)$ and $L$. 

While linear mapping is simple, it illustrates the rationale of stitching, which is to convert representations from one space to another. Recently, \citeauthor{din_jump_2023} \shortcite{din_jump_2023} proposed using linear mapping to replace the complicated transformer and convert between layers of hidden representations. The widely used LoRA method adopts a similar rationale in replacing complicated transformations with simple matrix multiplications \cite{hu_lora_2022}. In distillation for pre-training, hidden state vectors are similarly aligned between teacher and student. Our approach differs from theirs in that we connect the representation spaces using adjacent sentence pairs. 

\subsubsection{Loss Stitching.}

While it is possible to derive the mapper through a separate training process and corpus, we design a simple stitching framework instead, as in Figure \ref{fig4}. It combines the objective function of improving classification with the objective function of improving mapping, which we call Loss Stitching. In the figure, we use a dashed line to illustrate that the mapper is determined by objective 2. In implementation, it is done through neural networks’ backward propagation. In this way, the mapping can be abbreviated as: 
\begin{equation}
B(s_k)^*=LS(s_{k-1},s_{k-2})\cdot G(s_{k-1})
\end{equation}
where $LS(s_{k-1},s_{k-2})$ is just an annotation. It is determined by all sentence pairs in training rather than just $s_{k-1}$ and $s_{k-2}$ in each round. Objective function 2 can be defined on the MSELoss between the two representations (or CosineEmbeddingLoss as in distillation).
 
An obvious extension of linear mapping is to include more layers of neural networks and nonlinear activation functions in the mapper. In our experiments, we apply a $Tanh$ function between two linear layers for better performance. Meanwhile, using our Loss Stitching framework, we can choose to update the parameters of the entire or a part of neural network through fine-tuning. 

While nonlinear mapping and fine-tuning are expected to improve performance, they also raise concerns that the high-dimensional model will remember training sentence pairs $(s_{k-1}, s_k)$. To prevent this, we include a masking step that randomly masks the input of BERT in training. Masking is not needed in the evaluation/prediction stage.

\subsubsection{Attention Stitching.}

\begin{figure}[t]
\centering
\includegraphics[width=0.8\columnwidth]{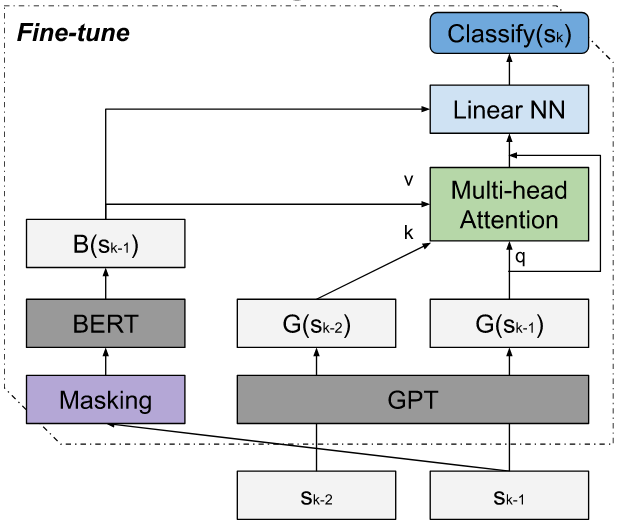} 
\caption{Attention Stitching}
\label{fig5}
\end{figure}

The Loss Stitching framework is simple but differs from the popular stacking and scaling approach in transformer-based LLMs. In the literature, attention modules have been used to fuse multimodal data \cite{liu_attention-based_2023}. Even though our setting has only text data, the fact that we have different sentences in a model also makes it possible to use Attention Stitching, which we introduce in Figure \ref{fig5}.

One may be aware that Loss Stitching essentially looks for a weight $LS$ multiplied with $G(s_{k-1})$. This is close to the attention mechanism:
\begin{equation}
attn(Q, K, V)=softmax(Q\cdot K^T/\sqrt{d_k})\cdot V
\end{equation}
In the attention framework, $Q$ in one space is converted to the space of $V$, where the conversion depends on the similarity of $Q$ and $K$ ($K$ corresponds to $V$ but is in the $Q$ space). In other words, if we take $G(s_{k-1})$ as $Q$, $G(s_{k-2})$ as $K$, and $B(s_{k-1})$ as $V$, we expect to see that an attention mechanism can drive the output of GPT toward BERT space based on the relevance between $B(s_{k-1})$ and $G(s_{k-2})$. If we represent $softmax(Q\cdot K^T/\sqrt{d_k})$ as $AS(Q, K)$, we have Attention Stitching as:
\begin{equation}
B(s_k)^*=AS(G(s_{k-1}), G(s_{k-2}))\cdot B(s_{k-1})
\end{equation}
where $AS(G(s_{k-1}), G(s_{k-2}))$ is also an annotation, which is determined by all training data. After that, we expect the well-established transformer framework to figure out the proper parameters that complete the mapping. It should be noted that even though equations 3 and 5 are in the same shape, they optimize different objectives.

\section{Experiments and Results}

\subsection{Dataset}

In this paper, we employ a large public dataset PUBMED-RCT \cite{dernoncourt_pubmed_2017} for our experiments. It contains 20K biomedical abstracts from PubMed, with sentences classified as BACKGROUND, OBJECTIVE, METHOD, RESULT, and CONCLUSION. The scale of the dataset allows us to capture the possibilities of embedding mapping between adjacent sentences. We use the preprocessed version of this dataset by \citeauthor{jin_hierarchical_2018} \shortcite{jin_hierarchical_2018}, which is available to the public on HuggingFace. 

\begin{table}[t]
\centering
\small 
\begin{tabular}{l|r|r|r|c|r}

\hline
									&  \multicolumn{1}{c|}{\textbf{N}}	& \textbf{Mean} 	& \textbf{S.D.}	& \textbf{Min} 	& \textbf{Max} \\
\hline
		Sent./Doc	& 20,685	& 11.40	& 3.44	& 1	& 31\\
		Sent./Background	& 10,161	& 2.40	& 1.26	& 1	& 12\\
		Sent./Objective	& 12,594	& 1.47	& 0.84	& 1	& 10\\
		Sent./Method	& 19,330	& 4.09	& 1.94	& 1	& 17\\
		Sent./Result	& 18,767	& 4.13	& 1.80	& 1	& 18\\
		Sent./Conclusion	& 19,772	& 1.83	& 0.91	& 1	& 11\\
		Word/Sentence 	& 235,892	& 26.69	& 15.31	& 1	& 296\\
\hline
\end{tabular}
\caption{Data Summary}
\label{table1}
\end{table}

Table \ref{table1} shows the summary statistics of the dataset. As we can see, on average, each document has about 11 sentences; each sentence has about 27 words. Half of the abstracts have all sections. However, almost every abstract has method, result, and conclusion sections. Generally, each section has about two to four sentences.

\subsection{Experiment Procedure}

The experiments are conducted on a DELL workstation with two Xeon E5-2630 CPUs, 128G memory, and a Gigabyte GeForce GTX TITAN X GPU. Due to resource limitations, we conduct experiments on pre-trained models with a relatively smaller number of parameters, including ``bert-base-uncased,'' ``gpt-2,'' and ``facebook/bart-base,'' which are publicly available from HuggingFace. 

\subsubsection{Feature-based vs. Fine-tuning.}

In experiment I, we experiment with both the feature-based approach and the fine-tuning approach. The feature-based approach uses the outputs of LLMs as features for traditional classification models. In our experiments, we use the BERT model and follow the routine to tokenize the sentence feed to the BERT model. Then, we use the embedding corresponding to the [CLS] token as features for the XGboost model in the sci-kit learn package for classification.

The major approach we take in the paper is fine-tuning, which adds a classification head upon a pre-trained model and allows the parameters of the pre-trained model to adapt to a specific task. For the BERT models, we apply classification head on the embedding of the [CLS] token of sentences. For the GPT and BART models, we use embedding of the [EOS] token for classification. In the scenarios where multiple sentences are used, these embeddings are generated for each sentence and combined before feeding them to a linear layer for classification.

\subsubsection{LASI Models.}

Based on the fine-tuning framework, we tokenize the input sentence(s) and transform them to embedding(s). Within fine-tuning, the classification head can be tweaked to improve performance. The default setting of sequence classification uses a linear layer. In our setup, we try to maintain this simplicity. We define the loss function as a regular sequence classifier using CrossEntropyLoss (except for Loss Stitching). In summary, we compare the performances of the baselines/proposed methods below for LASI:
\begin{itemize}
    \item BERT, GPT, and BART directly transform the input to the embeddings of the three models supplemented with a linear layer for classification. In the setting where multiple sentences are used as input, we convert them to embeddings separately and concatenate the embeddings before feeding them to the linear layer for classification. 
    \item BERT(GPT) first generates 50 new tokens using GPT for each sentence. Then, the two sentences are fed to BERT as a pair to generate the embedding, which is later sent to the linear layer for classification. 
		\item GPT+BERT first transforms the input to the embeddings of the two models and then concatenates the embeddings before feeding to the linear layer for classification. 
		\item GBLS (GPT BERT Loss Stitching) implements Loss Stitching based on BERT and GPT. It takes the input to the two models to generate their embeddings. Then, as shown in Figure \ref{fig4}, the GPT embedding is transformed through a $Tanh$ function between two linear layers (decided through small-scale comparison with a linear layer) before concatenating it with the BERT embedding and feeding to the linear layer for classification. The loss function contains the sequence classification CrossEntropyLoss and MSELoss for GPT embedding transformation (with a weight of 0.05 after small-scale experiments).
		\item GBAS (GPT BERT Attention Stitching) implements the Attention Stitching on BERT and GPT. It takes the two types of embeddings through a multihead attention (8 heads, decided arbitrarily), as illustrated in Figure \ref{fig5}, before feeding to the linear layer for classification. 
\end{itemize}
The codes for these methods are posted on GitHub\footnote{https://github.com/Julian-JJ/LLM\_Look\_Ahead\_Classification} for others to reproduce our experiments.

It should be noted that the performance of LLMs depends on many factors, such as the pre-training corpus, tokenizer, fine-tuning procedure and data. Due to resource limitations, this study mainly works on the fine-tuning stage for improvements. In our setup, we fine-tune the entire model and do not optimize efficiency, such as by using LoRA. Lastly, we mainly focus on the sentence content to help us make predictions and do not exploit the label interdependence of adjacent sentences, as CRF does.

\subsubsection{Evaluation.}

We use the dataset’s original split of train, validation, and test data. We use the train and validation data to fine-tune parameters for 10 epochs. We choose the epoch with the best accuracy as the final model (which often appears in the first 5 epochs) to evaluate the testing data. In the experiments, hidden vectors are in 768 dimensions. We truncate and pad sentences to 100 tokens. In the experiments, we set the learning rate as $2\cdot{10}^{-5}$ and set weight decay as 0.01 without much tuning. In GBLS and GBAS, we experiment with masking and choose masking 10\% tokens for better performance over two other options 0\% and 20\%. 

We conduct the task as a multi-class classification problem, where each sentence is classified into one of the multiple labels. We compare the prediction with the actual label to calculate accuracy and weight-averaged precision/recall/F1-measure (on the number of instances of each category). Accuracy represents the percentage of correct classifications. Precision is the ratio of instances that the model-specified label is correct. Recall is the ratio of instances that the sentences of a certain category are correctly classified. F1 score is the harmonic mean of precision and recall to balance the two measures. These are widely used measures to evaluate multi-class classification tasks. 

\subsection{Experiment I: Problem Validity}

In the first experiment, we compare SI and LASI using both feature-based and fine-tuning methods. The differences between them are: 1) learning head: XGBoost vs. linear neural network; and 2) fixed vs. adjustable BERT parameters. 

The results are reported in Table \ref{table2}. First, we notice that SI, which seems to be a resolved problem, can be effectively tackled by BERT with 80-90\% accuracy. LASI is a more challenging task. The performance of BERT on LASI is 10-15\% less than classic SI, which creates room for us to improve. This performance difference is mainly caused by the fact that LASI misses the focal sentence to classify, causing the model to make inferences from the context, which is significantly more challenging.

Second, as expected, the fine-tuning method provides us with better performance. (The default output of scikit-learn is two digits, which does not hinder our comparison here.) Since a 1-layer linear layer is less powerful than XGBoost, the performance improvement is mainly due to fine-tuning. We choose fine-tuning as the major approach for further experiments.

\begin{table}[t]
\centering
\small 
\begin{tabular}{l|l|c|c|c|c}
\hline
		\multicolumn{2}{c|}{\textbf{Feature-based}} 	& \textbf{Acc}	& \textbf{P}	& \textbf{R}	& \textbf{F1}\\
\hline
		SI	& $s_k$	& 80\%	& 80\%	& 80\%	& 79\%\\
			& $s_{k-1}s_k$	& 84\%	& 84\%	& 84\%	& 84\%\\
			& $s_{k-1}s_ks_{k+1}$	& 86\%	& 86\%	& 86\%	& 86\%\\
		LASI	& $s_{k-1}$	& \textbf{68\%}	& \textbf{67\%}	& \textbf{68\%}	& \textbf{67\%}\\
			& $s_{k-2}s_{k-1}$	& 73\%	& 72\%	& 73\%	& 72\%\\
\hline
		\multicolumn{2}{c|}{\textbf{Fine-tuning}}	& \textbf{Acc}	& \textbf{P}	& \textbf{R}	& \textbf{F1}\\
\hline
		SI	& $s_k$	& 86.4\%	& 86.5\%	& 86.4\%	& 86.4\%\\
			& $s_{k-1}s_k$	& 90.5\%	& 90.5\%	& 90.5\%	& 90.4\%\\
			& $s_{k-1}s_ks_{k+1}$	& 91.6\%	& 91.5\%	& 91.6\%	& 91.5\%\\
		LASI	& $s_{k-1}$	& \textbf{74.0\%}	& \textbf{73.7\%}	& \textbf{74.0\%}	& \textbf{73.1\%}\\
			& $s_{k-2}s_{k-1}$	& 77.2\%	& 76.9\%	& 77.2\%	& 76.6\%\\
\hline
\end{tabular}
\caption{Results on BERT-based Baselines}
\label{table2}
\end{table}

Third, we notice that combining the last two sentences in LASI does not make much difference as compared to using the last sentence. Thus, we use $s_{k-1}$ to predict the label of $s_k$ in the following experiments.

\subsection{Experiment II: Model Comparison on LASI}

Table \ref{table3} reports the results of the proposed models, where we only use $s_{k-1}$ to predict the label of $s_k$.

First, GPT does not show an advantage over BERT, which may be due to its limited ability to capture contextual information. Meanwhile, the straightforward ways to combine GPT and BERT, such as using GPT-generated content as BERT input, BERT(GPT), or directly combining GPT and BERT, BERT+GPT, do not generate better results than BERT alone. These simple approaches cannot solve LASI. More advanced models are needed to exploit the power of BERT and GPT for LASI. 

Second, we observe that complicated models, including BART and our proposed models, show performance advantages over BERT. In particular, our proposed methods, GBLS and GBAS, have the best performances. We believe it is their ability to drift GPT toward BERT representation that improves the model’s ability to perform LASI. 

In GBLS, we apply $Tanh$ activation between two linear layers as a mapper. Small experiments show that such a design can make the MSE of the two representations to 0.07 (i.e., RMSE close to 0.26). This is about 10\% of the BERT representation. Even with this level of error, the model still generates satisfying results.

GBAS shows about 1\% increase in both accuracy and F1 measure over BERT.\footnote{Someone may be concerned that GBAS uses $s_{k-2}$ in the prediction stage (which GBLS does not). To address this concern, we create another model incorporating $B(s_{k-2})$ into GBAS. It still generates about 1\% improvement in accuracy and F1 measure over the last row of Table \ref{table2}.}  With that being said, it should be noted that GBAS highly relies on the neural network to figure out the parameters, and we have less control over it than GBLS, which may be necessary for some applications. 

It should also be noted that while using attention modules, Attention Stitching is fundamentally different from self-attention, such as encoder-decoder attention, since it is based on the assumption that the GPT representation and the BERT representation on adjacent sentences are related. Given its performance in LASI, it would be worthwhile to study the value of Attention Stitching in other applications. 

\begin{table}[t]
\centering
\begin{tabular}{l|c|c|c|c}
\hline
		\textbf{$s_{k-1}$}	& \textbf{Acc}	& \textbf{P}	& \textbf{R}	& \textbf{F1}\\
\hline
		BERT	& 74.0\%	& 73.7\%	& 74.0\%	& 73.1\%\\
		GPT	& 74.0\%	& 74.0\%	& 74.0\%	& 73.1\%\\
		BERT(GPT)	& 73.7\%	& 73.3\%	& 73.7\%	& 73.4\%\\
		GPT+BERT	& 73.8\%	& 74.0\%	& 73.8\%	& 73.0\%\\
		BART	& 74.3\%	& 73.7\%	& 74.3\%	& 73.6\%\\
		GBLS	& 74.7\%	& 74.3\%	& 74.7\%	& 74.1\%\\
		GBAS	& \textbf{74.8\%}	& \textbf{74.5\%}	& \textbf{74.8\%}	& \textbf{74.3\%}\\
\hline
\end{tabular}
\caption{Results of Proposed Models}
\label{table3}
\end{table}

\subsection{Experiment III: Model Robustness }

Then, we tweak the task to inspect whether introducing noise into the developing text will affect the performance of models. The tweaks include:
\begin{itemize}
    \item Randomly removing 1 or 2 words from the sentence $s_{k-1}$, which mimics the setting where a sentence is not carefully written. 
    \item Adding the next 1 or 2 words from sentence $s_k$, which mimics the setting where a couple of words are written. 
\end{itemize}
In these two settings, we only tweak the testing data to add noise and do not change the validation and training data, mimicking real noise scenarios.

\begin{figure}[b]
\centering
\includegraphics[width=0.9\columnwidth]{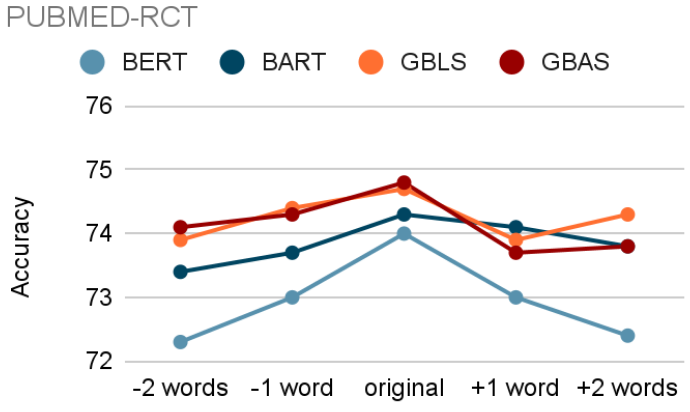} 
\caption{Performance on Tweaked Settings}
\label{fig6}
\end{figure}

Table \ref{table4} reports the experimental results of these various settings, which are also illustrated in Figure \ref{fig6}. Noting the correlation between performance measures, we only report the accuracy results here. Other performances are highly similar. As we can see, the advantage of stitching and BART models over BERT increases when the sentences get noisier. As we know, noise is common in a generative AI context. Targeting this application, stitching models are more robust than the BERT model.  

\begin{table}[t]
\centering
\begin{tabular}{l|c|c|c|c}
\hline
		\textbf{$s_{k-1}$}	& \textbf{Acc}	& \textbf{P}	& \textbf{R}	& \textbf{F1}\\
\hline
		-2 words	& 72.3\%	& 73.4\%	& 73.9\%	& 74.1\%\\
		-1 word	& 73.0\%	& 73.7\%	& 74.4\%	& 74.3\%\\
		original	& 74.0\%	& 74.3\%	& 74.7\%	& 74.8\%\\
		+1 word	& 73.0\%	& 74.1\%	& 73.9\%	& 73.7\%\\
		+2 words	& 72.4\%	& 73.8\%	& 74.3\%	& 73.8\%\\
\hline
\end{tabular}
\caption{Results on Tweaked Settings}
\label{table4}
\end{table}

\subsection{Discussion}

Through the experiments, we learn a few lessons:
\begin{itemize}
    \item LASI is more difficult to tackle than classic SI beacuse it is missing the focal text to classify.
    \item Our proposed stitching framework provides a good solution that combines BERT (bidirectional autoencoder) with GPT (unidirectional autoregressive models) to tackle the LASI task. 
    \item Stitching is effective when there is noise in the data, which is common in generative AI. 
\end{itemize}

While our main experiments are conducted on BERT and GPT, our stitching framework is not limited to these two models. In the past, many pre-trained LLMs were created (using different corpora, different languages, or for different purposes), which cost a lot of computational power. The stitching approach brings a new perspective to LLM modeling and allows us to use these various pre-trained LLMs even though their individual performance may not be outstanding on a task. This could be a lightweight approach to leverage existing efforts on LLMs.

\section{Related Works}

\subsection{Section Identification}

SI is a text classification task \cite{minaee_deep_2021}. Several papers have reviewed this problem, such as \citeauthor{zhou_feature_2020} \shortcite{zhou_feature_2020} and \citeauthor{ma_enhancing_2022} \shortcite{ma_enhancing_2022}. In literature, a number of studies take a feature-based approach and derive linguistic features (e.g., bag-of-words, part-of-speech, lexicon, and N-gram), structural features (e.g., heading, sentence location, and sentence length), semantic features (e.g., synonyms, and hypernyms), and citation features to address this problem \cite{hassanzadeh_identifying_2014, zhou_feature_2020}. Since there exist highly indicative features, such as section headings, this approach was very effective. Nevertheless, we are more interested in generalizable approaches that exploit the semantics of sentences, where deep learning provides another effective venue \cite{minaee_deep_2021}. Below, we mainly review recent SI studies that use deep learning. 

By using a deep learning approach, one can build customized neural network structures. When working with textual data, two commonly used neural network structures are LSTM and Bi-LSTM. For example, \citeauthor{jin_hierarchical_2018} \shortcite{jin_hierarchical_2018} employed BiLSTM+CRF after sentence embedding to classify the rhetorical structure of medical scientific documents. Upon their framework, \citeauthor{brack_sequential_2024} \shortcite{brack_sequential_2024} replaced the pre-trained embedding to improve scientific document processing ability. \citeauthor{zhou_feature_2020} \shortcite{zhou_feature_2020} employed BiLSTM to model within sentence structure and used a dense layer to model across sentence relations for this task. \citeauthor{goncalves_deep_2020} \shortcite{goncalves_deep_2020} built a structure with CNN and GRU for this task. Hierarchical attention networks were also used in literature \cite{ma_enhancing_2022}.

Recent advances in pre-trained LLMs provide us with a new way to capture sentence information for SI. Due to the powerful ability of large-scale models to capture the interdependency of sentences (which comes from the pre-training materials), large-scale models can achieve quite good performances. For example, \citeauthor{cohan_pretrained_2019} \shortcite{cohan_pretrained_2019} employed BERT to process multiple sentences within a paragraph and then feed them into a dense layer for classification. This simple model can outperform many traditional baselines \cite{jin_hierarchical_2018}. Such an approach to directly apply BERT on SI can be seen in various application studies \cite{de_la_iglesia_open_2023,gray_classifying_2023}. \citeauthor{yu_masked_2019} \shortcite{yu_masked_2019} used both the focal sentence and the rest of the document as inputs of BERT to better capture the context of the focal sentence and enhance performance. \citeauthor{jiang_biomedical_2023} \shortcite{jiang_biomedical_2023} took a reading comprehension approach to find the starting and ending points of a section of sentences in biomedical abstracts. \citeauthor{zhou_self-supervised_2021} \shortcite{zhou_self-supervised_2021} argued that by regularizing the classifier with self-supervised tasks, such as masked token prediction and sentence augmentation type prediction, the main task of classification can be improved. In recent years, prompt engineering has also been introduced to this domain \cite{hu_towards_2023}, which is less relevant to this study. 

As we can see from the literature, while tremendous effort has been exerted to leverage deep learning for text classification and specifically the SI task, these approaches are significantly different from our proposed stitching approach. 

\subsection{Developing Text and Generative AI}

In recent years, generative AI has seen significant advances. Generative AI models are able to perform a variety of tasks, such as divergent thinking tasks, in which AI outperforms the average human, if not the best humans \cite{koivisto_best_2023}. While generative AI is most well-known for dialogue generation and multimodal content generation (such as generating images based on textual prompts), it has also achieved significant attention for its long text generation \cite{liang_open-ended_2023}, which often uses autoregressive models, such as GPT or BART. While these models have gained popularity and demonstrated effectiveness, their autoregressive nature has resulted in inefficiencies.

In long text generation, another challenge is to maintain a logic or a coherent storyline within the generated text \cite{goldfarb-tarrant_content_2020}. A common solution is to add a content planning stage and design plots to direct the generation model \cite{yao_plan-and-write_2019}. To a certain extent, such plots are similar to structures of scientific and professional documents. In practice, it is also common to interactively involve humans in the text generation process so that deviation from the logic can be corrected. Our study considers the needs and noises that may appear in this approach to explore LASI.  

\section{Conclusion}

In the paper, we tackle the LASI problem as a particular case of a class of look ahead text understanding problems, such as look ahead sentiment classification and look ahead topic classification, which is necessary for analyzing social media conversations. We propose an LLM stitching approach to leverage BERT and GPT to address the LASI problem. We find that the approach outperforms existing approaches, especially in contexts with noise.

While our main experiments tackle LASI and are conducted on science abstracts due to data availability and our expertise, we argue that look ahead text understanding is generic to different applications. For example, the approach can be directly used in social media, such as web forums where users reply in turns in a thread. The forum structure generates a context of developing text, where one may want to predict the direction of a conversation (such as topics or sentiments). While our experiments on scientific documents provide a good baseline, experiments should be conducted on social media corpora to explore the directions of conversations. Our LLM stitching framework can also be applied to these problems to evaluate their performances.

In addition to social media applications, look ahead text understanding may also be useful in the development of other AI systems. For example, in generative AI, co-pilots can search tools/software/reports on the disk while projecting that the user will need some of these items in follow-up communications/document writing. This may significantly impact the user experience or outcomes and bring a positive surplus to the world.

\subsection{Ethical Concerns}

As techniques advance, their ethical implications should not be ignored. While our experiments are immune from this challenge due to a special test bed, look ahead text understanding in general may lead to ethical concerns when applied to social media and other content involving individuals' data and lives. On the one hand, the fact that this prediction is made based on one's historical expressions may lead to privacy concerns and may require consent. In addition, in a context of accurate prediction abilities, people may have a feeling that they are being monitored, which may lead to resistance to this functionality. On the other hand, it is necessary to discuss the implications of ``making inferences based on words that are not being said (yet).'' If the technology is misused, it is possible that individuals are misunderstood even before they express their views, which may hinder conversation effectiveness. It is also possible that the AI's prediction is wrong or that the person changed his/her opinion/topic (aided by another look ahead text understanding tool). In such scenarios, unexpected results may emerge. It would be necessary for future research to account for these concerns in the development of generative AI or look ahead text understanding tools.
		
\subsection{Limitations}
The paper has its limitations that could be addressed in the future. First, our experiments are based on one dataset focusing on the LASI task. As we discussed, our problem framing can be generalized to other look ahead text understanding tasks, such as look ahead sentiment classification. Future research can test the proposed model on such tasks if they involve datasets with sentence-level or paragraph-level sentiment coding. Second, our evaluation is compared with established baselines in SI. There may exist other approaches that advance the performance of SI, which can be included in baselines. It is also possible to examine whether the proposed approach can be augmented with existing state-of-the-art methods to improve its performance further. These efforts could offer a more nuanced understanding of the proposed approach's advantages and limitations. Third, more careful parameter tuning could be conducted, which may further improve model performance. Fourth, more theoretical analyses should be conducted on stitching methods, including but not limited to efficiency, generalizability, and applicability. 

\section{Acknowledgments}
The work described in this paper was partially supported by the Research Grants Council of the Hong Kong Special Administrative Region, China [GRF 11500519], InnoHK initiative, The Government of the HKSAR, and Laboratory for AI-Powered Financial Technologies. 

JJJ conducted the coding and experiments, finished the data analysis, and drafted related parts of the paper. XL coined the research idea, directed the experiments, and wrapped up the paper. 

\bigskip
\bibliography{lasi}
\end{document}